\documentclass[letterpaper]{article} 
\usepackage{aaai25}  
\usepackage{times}  
\usepackage{helvet}  
\usepackage{courier}  
\usepackage[hyphens]{url}  
\usepackage{graphicx} 
\usepackage{natbib}  
\usepackage{caption} 
\usepackage{algorithm}
\usepackage{algorithmic}
\usepackage{url}  
\usepackage{booktabs}
\usepackage{amsfonts}  
\usepackage{nicefrac} 
\usepackage{microtype}   
\usepackage{fancyhdr}      
\usepackage{bm}
\usepackage{amsmath}
\usepackage{algorithmic}
\usepackage{algorithm}
\usepackage{multirow}
\usepackage{graphicx}
\usepackage{subfigure}
\usepackage{amsthm}
\usepackage{multicol}
\usepackage{pdfpages}
\usepackage{makecell}
\usepackage{xspace}
\usepackage{float}
\usepackage{threeparttable}
\usepackage{soul}
\usepackage{array}
\usepackage{amsthm,amsmath,amssymb}
\usepackage{dsfont}
\usepackage{pifont}
\usepackage{amssymb}
\usepackage{caption}
\usepackage{graphicx, subfig}

\usepackage{newfloat}
\usepackage{listings}
\DeclareCaptionStyle{ruled}{labelfont=normalfont,labelsep=colon,strut=off} 
\floatstyle{ruled}
\newfloat{listing}{tb}{lst}{}
\floatname{listing}{Listing}
\title{Dense Audio-Visual Event Localization under\\ Cross-Modal Consistency and Multi-Temporal Granularity Collaboration}
\author{
    Ziheng Zhou\textsuperscript{\rm 1},
    Jinxing Zhou\textsuperscript{\rm 1}\thanks{Corresponding authors},
    Wei Qian\textsuperscript{\rm 1},
    Shengeng Tang\textsuperscript{\rm 1},
    Xiaojun Chang\textsuperscript{\rm 2,3},
    Dan Guo\textsuperscript{\rm 1}\textsuperscript{$\ast$}
    \\
}
\affiliations{
    \textsuperscript{\rm 1}School of Computer Science and Information Engineering, Hefei University of Technology\\
    \textsuperscript{\rm 2}University of Science and Technology of China\\
     \textsuperscript{\rm 3}Mohamed Bin Zayed University of Artificial Intelligence\\
    zzhhfut@gmail.com, zhoujxhfut@gmail.com, guodan@hfut.edu.cn\\
}

\usepackage{bibentry}

\makeatletter
\DeclareRobustCommand\onedot{\futurelet\@let@token\@onedot}
\def\@onedot{\ifx\@let@token.\else.\null\fi\xspace}

\def\ie{\emph{i.e}\onedot} 
 
\def\etc{\emph{etc}\onedot} 
 
\def\etal{\emph{et al}\onedot}
\makeatother

\begin{document}

\maketitle

\begin{abstract}
In the field of audio-visual learning, most research tasks focus exclusively on short videos. This paper focuses on the more practical Dense Audio-Visual Event Localization (DAVEL) task, advancing audio-visual scene understanding for longer, {untrimmed} videos. This task seeks to identify and temporally pinpoint all events simultaneously occurring in both audio and visual streams. Typically, each video encompasses dense events of multiple classes, which may overlap on the timeline, each exhibiting varied durations.
Given these challenges, effectively exploiting the audio-visual relations and the temporal features encoded at various granularities becomes crucial. To address these challenges, we introduce a novel \ul{CC}Net, comprising two core modules: the Cross-Modal Consistency \ul{C}ollaboration (CMCC) and the Multi-Temporal Granularity \ul{C}ollaboration (MTGC). Specifically, the CMCC module contains two branches: a cross-modal interaction branch and a temporal consistency-gated branch.
The former branch facilitates the aggregation of consistent event semantics across modalities through the encoding of audio-visual relations, while the latter branch guides one modality's focus to pivotal event-relevant temporal areas as discerned in the other modality.
The MTGC module includes a coarse-to-fine collaboration block and a fine-to-coarse collaboration block, providing bidirectional support among coarse- and fine-grained temporal features. Extensive experiments on the UnAV-100 dataset validate our module design, resulting in a new state-of-the-art performance in dense audio-visual event localization. The code is available at \url{https://github.com/zzhhfut/CCNet-AAAI2025}.
\end{abstract}

\section{Introduction}
Hearing and vision are two crucial senses for humans in perceiving their surroundings.
Within the research community, recent years have seen a surge of interest in the joint exploration and comprehension of audio and visual signals, giving rise to numerous audio-visual learning tasks.
These include audio-visual event localization~\cite{tian2018audio,zhou2024towards} and video parsing~\cite{tian2020unified,zhou2023improving,zhou2024label,zhou2024vaplan,gao2023collecting,zhao2024mmcse}, sound source localization~\cite{senocak2021TAPMI,hu2019deep,qian2020multiple} and segmentation~\cite{zhou2022avs,zhou2023avss,mao2023multimodal,liu2023audio,li2023catr,guo2023audio}, audio-visual question answering~\cite{lao2023coca,li2022learning,yang2022avqa,li2024object,li2024post,li2023progressive} and captioning~\cite{tian2018attempt,iashin2020better,mao2024tavgbench,shen2023fine}, \etc.
However, the majority of these tasks have predominantly focused on trimmed videos of short durations, commonly 5s or 10s.

\begin{figure}[t]
  \centering
  \includegraphics[width=\linewidth]{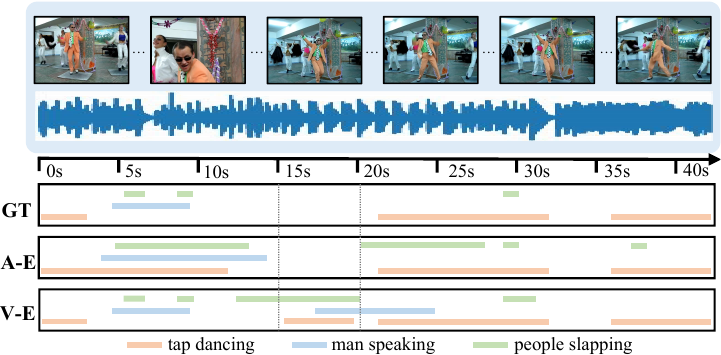}
  \caption{\textbf{Illustration of the Dense Audio-Visual Event Localization (DAVEL) task.} 
  The DAVEL task requires temporally localizing the events that occur simultaneously in both audio and visual tracks of untrimmed videos. These dense events may overlap on the timeline and vary in duration. 
  ``GT” denotes the ground truth for audio-visual events, which are the intersection of audio events (``A-E”) and visual events (``V-E”).}
  \label{fig:introduction}
\end{figure}

In this paper, we focus on a more realistic task termed Dense Audio-Visual Event Localization (DAVEL)~\cite{geng2023dense}, which concentrates on the scene understanding of long, \textit{untrimmed} audible videos.
Notably, the DAVEL can be considered an extension of the existing audio-visual event localization (AVEL)~\cite{tian2018audio} task.
Both tasks aim to identify and temporally localize the audio-visual events, events occurring in both audio and visual tracks.
The key difference is that the AVEL focuses on short videos (fixed at 10s) where each video contains only a specific event class.
In contrast, as illustrated in Fig.~\ref{fig:introduction}, the DAVEL task addresses \textit{untrimmed} videos, with an average length of 42.1s for the official UnAV-100~\cite{geng2023dense} dataset. Furthermore, each video usually contains \textit{dense} events, and events from multiple classes can temporally overlap, indicating the event's co-occurrence. 
More importantly, the AVEL is formulated as a \textit{segment-level classification} problem, while the DAVEL seeks to precisely regress the start and end timestamps of each detected event (\textit{frame-level regression}).
This task difference renders prior excellent works in the AVEL task inapplicable to the DAVEL task.
We will provide more introduction and discussions about this in the Related Work section. 

Here, we highlight three main characteristics of the studied DAVEL task.
\textbf{(C1) Cross-modal event-consistency.} The target audio-visual events are the \textit{intersection} between audio events and visual events. In other words, a DAVEL model aims to capture event semantics \textit{shared} between audio and visual modalities;
\textbf{(C2) Cross-modal temporal consistency.} Not all temporal segments contain audio or visual events, indicating that some segments may harbor background noise or other event-irrelevant information. For instance, as depicted in Fig.~\ref{fig:introduction}, there are no \textit{audio} events between 15s$\sim$20s. 
As a result, the ground truth contains no \textit{audio-visual} events regardless of the events present in the \textit{visual} track during this temporal period.
This decision is guided by the aforementioned intersection operation defined for audio-visual events.
For one modality, the model should also focus on key temporal regions identified in the other modality; \textbf{(C3) Event duration inconsistency.} As presented in Fig.~\ref{fig:introduction}, each event may span various temporal windows. Considering interactions among audiovisual features at different temporal granularities would benefit the model.

\begin{figure*}[t]
    \centering
\includegraphics[width=\textwidth]{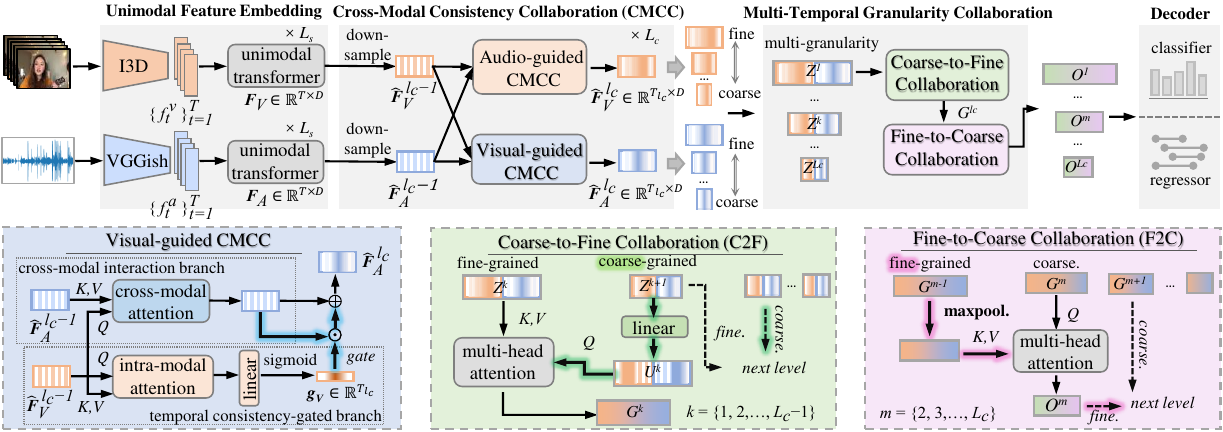}
    \caption{\textbf{The pipeline of our CCNet for the dense audio-visual event localization task.}}
    \label{fig:framework}
\end{figure*}

Motivated by these observations, we propose a new \ul{CC}Net for DAVEL task, comprising two core modules: the \textbf{Cross-Modal Consistency \ul{C}ollaboration (CMCC)} and the \textbf{Multi-Temporal Granularity \ul{C}ollaboration (MTGC)}.
The details of each module are illustrated in Fig.~\ref{fig:framework}.
Specifically, {the CMCC module} draws its design from the analyses of the first two characteristics outlined above.
The CMCC includes two branches: a cross-modal interaction branch and a temporal consistency-gated branch.
The former branch encodes audio-visual relations through multi-head attention~\cite{vaswani2017attention} mechanism, enabling each modality to aggregate \textit{consistent event semantics} from the counterpart modality \textbf{(C1)}.
The latter, a temporal consistency-gated branch, initiates by encoding unimodal relations within one modality via self-attention. 
Then, the encoded feature is utilized to learn a temporal weight vector highlighting \textit{key event-related temporal regions}, subsequently serving as a temporal-wise gate to regularize the feature of the other modality \textbf{(C2)}.
It is noteworthy that multiple stacked CMCC modules are utilized in our network, with a feature downsampling operation implemented at the outset of each CMCC module, producing features at multiple temporal scales.
{Regarding the MTGC module}, it encompasses a Coarse-to-Fine collaboration (C2F) and a Fine-to-Coarse collaboration (F2C) block. 
In general, temporal features with relatively high downsampling rate are considered coarse-grained, while those with a lower downsampling rate are deemed fine-grained. 
Coarse-grained features prove advantageous in delineating coarse temporal regions of events occurring in the video, whereas fine-grained features aid in predicting precise temporal boundaries of events.
The C2F and F2C blocks facilitate \textit{bidirectional collaboration} between the coarse- and fine-grained features \textit{across multiple temporal granularities}, benefiting the localization of events with varied durations \textbf{(C3)}.

Extensive experimental results on the UnAV-100 dataset demonstrate the effectiveness and superiority of our method.
Our main contributions can be summarized as follows:
\begin{itemize}
\item We identify and analyze three key characteristics of the DAVEL task, which leads to a new CCNet approach comprising several simple yet highly effective modules.
\item We design a Cross-Modal Consistency Collaboration module, which incorporates both a cross-modal interaction branch and a temporal consistency-gated branch, ensuring superior audio-visual representation embedding.
\item We introduce a Multi-Temporal Granularity Collaboration module, which features coarse-to-fine and fine-to-coarse collaboration blocks, enabling the model to utilize temporal features bidirectionally across various granularities.
\item Our method achieves a new state-of-the-art on the UnAV-100 dataset, surpassing the previous baseline in mAP metrics across multiple tIoU thresholds and exhibiting superior performance in localizing events of varied durations.
\end{itemize}

\section{Related Work}\label{sec:related_work}
\noindent\textbf{Audio-Visual Event Localization (AVEL)} task aims to identify the video segments containing a specific audio-visual event (both audible and visible) and classify its category. The pioneer work~\cite{tian2018audio} proposes a dual multimodal residual network to fuse audio and visual features.
Zhou~\etal~\cite{zhou2021positive,zhou2023contrastive} design a positive sample propagation module to select the most highly relevant audio-visual pairs for feature aggregation. To deal with audio-visual events existing in different temporal scales, Yu \etal~\cite{yu2021mm} constrain the audio-visual interaction in multiple fixed-size temporal windows.
Although those methods have achieved significant progress for the AVEL problem, they are designed for trimmed videos in a short duration and can only realize the \textit{segment-level event classification}. 
In contrast, the studied DAVEL task tackles long untrimmed videos and requires \textit{frame-level timestamp regression}, which can not be solved by those AVEL methods. 

\noindent\textbf{Audio-Visual Video Parsing (AVVP)} task~\cite{tian2020unified} aims to comprehensively localize the audio events, visual events, and audio-visual events.
Unlike the AVEL and the studied DAVEL tasks, AVVP task does not emphasize audio-visual alignment.
Moreover, the AVVP task is performed in a weakly supervised setting, where only the event label of the whole video is available for model training.
Some researchers focus on developing more effective methods for audio-visual feature interaction ~\cite{yu2021mm,lin2021exploring,jiang2022dhhn,zhou2024label}.
Others try to generate video-level~\cite{wu2021exploring,cheng2022joint} or segment-level pseudo labels~\cite{lai2023modality,zhou2024vaplan} via label denoising or by utilizing pretrained large-scale models for better model optimization.
However, these AVVP methods remain limited to short, trimmed videos, rendering them unsuitable for the DAVEL task.

\noindent\textbf{Dense Audio-Visual Event Localization (DAVEL)} 
task aims to temporally localize all the audio-visual events appearing in untrimmed videos, predicting the corresponding event categories and temporal boundaries. 
Notably, the DAVEL is a newly proposed research task.
The pioneering work~\cite{geng2023dense} serves as a strong baseline for comparison.
Specifically, the backbone of this baseline includes a cross-modal pyramid transformer encoder (CMPT) and a temporal dependency modeling (TDM) module. The CMPT is used to encode the audio-visual relations at different temporal scales, while the TDM is designed to model the dependencies of the temporal segments and the concurrent events in different categories. Although this strong baseline has achieved impressive localization results, it simply encodes the cross-modal interaction via vanilla transformer encoders and overlooks the cooperation among features of various temporal scales. In contrast, our method considers enhancing the cross-modal consistency and multi-temporal granularity collaboration.

\section{Methodology}
\subsection{Problem Formulation}
Given an audible video sequence, it is first divided into $T$ segments at equal intervals.
Let $\{V_t, A_t\}_{t=1}^T$ denote the visual and audio segment pairs, respectively. 
The ground truth for each video is given as $Y = \{y_n = (t_{s,n}, t_{e,n}, c_n)\}_{n=1}^N$, indicating there are $N$ audio-visual events in the video.
For the $n$-th event, $t_{s,n}$ and $t_{e,n}$ are the start and end timestamps, respectively, and $c_n$ represents the event category, where $c_n \in \{1, ..., C\}$ ($C$ is the total class number of audio-visual events in the dataset).

The dense audio-visual event localization task is formulated as a sequence labeling and regression problem: for each timestamp $t$, the model needs to classify its event category and regress the distances from this moment to the event's start and end timestamps.
Consequently, the prediction is given as $\hat{Y} = \{ \hat{y}_t = (d_{s,t}, d_{e,t}, p(c_t)) \}_{t=1}^T$, where $p(c_t)\in \mathbb{R}^{1 \times C}$ is the event probability, $d_{s,t}$ and $d_{e,t}$ are the regressed onset and offset distances, respectively.
It is noteworthy that $d_{s,t}$ and $d_{e,t}$ are predicted only when an audio-visual event exists at moment $t$.
The final localization results can be obtained by post-processing the predictions as:
\begin{small}
\begin{equation}
c_t = \text{argmax}(p(c_t)), \quad t_{s,t}=t-d_{s,t}, 
\quad t_{e,t}=t+d_{e,t}.
\label{eq:problem_statement}
\end{equation}
\end{small}

\subsection{Framework Overview}
The overall pipeline of our framework is illustrated in Fig.~\ref{fig:framework}.
It consists of four main modules:
(1) At the initial \textbf{Unimodal Feature Embedding} module, video frames and audio signals are preprocessed, and their corresponding features are extracted using off-the-shelf pretrained convolutional neural networks.
The audio and visual features are further refined by encoding unimodal temporal relations.
(2) Subsequently, the proposed \textbf{Cross-Modal Consistency Collaboration (CMCC)} module generates audio and visual features at various temporal scales/granularities.
Each CMCC block consists of \textit{a cross-modal interaction branch} and \textit{a temporal consistency-gated branch}.
The former branch focuses on encoding cross-modal audio-visual relations to aggregate information on shared events in both modalities, while the latter branch ensures temporal consistency across audio and visual modalities, guiding the feature from one modality to focus on key temporal regions identified in the other modality.
(3) The updated audio and visual features are concatenated and transferred to the \textbf{Multi-Temporal Granularity Collaboration (MTGC)} module, designed to bolster interactions among features across different temporal granularities. 
The MTGC contains \textit{a Coarse-to-Fine (C2F)} and \textit{a Fine-to-Coarse (F2C) collaboration} block, allowing a bidirectional feature flow between coarse-grained and fine-grained temporal features. 
(4) Finally, the refined audiovisual features are forwarded to the \textbf{Decoder}, which uses a classification head to predict event probabilities and a regression head to determine the onset and offset distances.

\subsection{Unimodal Feature Embedding}
Given audio and visual components $\{A_t, V_t\}_{t=1}^T$, we extract their features following the baseline~\cite{geng2023dense}.
Specifically, the VGGish~\cite{hershey2017vggish} is utilized to extract the audio feature $\bm{a} =\{f_t^a\}_{t=1}^T \in \mathbb{R}^{T \times d_a}$.
For the visual feature extraction, the two-stream I3D~\cite{carreira2017quo} model is used, yielding the visual feature $\bm{v} = \{f_t^v\}_{t=1}^T \in \mathbb{R}^{T \times d_v}$.
Then, we use two convolutional layers to project the audio and visual features into the same embedding space, yielding $\bm{a}, \bm{v} \in \mathbb{R}^{T \times D}$, where $D$ is the feature dimension.
Note that the lengths of videos may vary, thus the feature vectors obtained by the pretrained models are cropped or padded to the max sequence length $T$.

Considering that audio or visual signals representing an event typically occur in consecutive timestamps, we further encode the uni-modal temporal relations via the self-attention mechanism.
This can be conveniently implemented by feeding the audio feature $\bm{a}$ or visual feature $\bm{v}$ into $L_s$ stacked Transformer~\cite{vaswani2017attention} blocks, with distinct parameters for each modality.

Through the aforementioned steps, we can obtain the unimodal audio and visual embeddings, denoted as $\bm{F}_A, \bm{F}_V \in \mathbb{R}^{T\times D}$, respectively.
Next, we consider facilitating the dense audio-visual event localization from two task-relevant perspectives: cross-modal consistency and multi-temporal granularity collaboration.
We elaborate on their design principles and detailed operations next.

\subsection{Cross-modal Consistency Collaboration}
The target audio-visual events in the video are both audible and visible.
Except for encoding the unimodal relations, one modality should also be aware of events within another modality and aggregate consistent event information from the other.
Given the audio and visual embeddings $\bm{F}_A, \bm{F}_V \in \mathbb{R}^{T\times D}$, we utilize \textbf{a Cross-Modal Interaction branch (CMI)} which encodes cross-modal relations through the multi-head attention (MHA)~\cite{vaswani2017attention}, followed by residual connection.
Specifically, the feature of one modality is used as the \textit{key} $\bm{\mathcal{K}}$ and \textit{value} $\bm{\mathcal{V}}$, and the feature of another modality is used as the \textit{query} $\bm{\mathcal{Q}}$, formulated as,
\begin{equation}\label{eq:MHA}
\begin{gathered}
\bm{\hat{F}}_{A} = \bm{F}_A + \text{MHA}(\bm{F}_V, \bm{F}_A, \bm{F}_A), \\
\bm{\hat{F}}_{V} = \bm{F}_V + \text{MHA}(\bm{F}_A, \bm{F}_V, \bm{F}_V), \\
\text{MHA}(\bm{\mathcal{Q}}, \bm{\mathcal{K}}, \bm{\mathcal{V}}) = \delta( \frac{\bm{\mathcal{Q}\bm{W}}^Q (\bm{\mathcal{K}}\bm{W}^K )^{\top}} {\sqrt{D}} ) \bm{\mathcal{V}}\bm{W}^V,
\end{gathered}
\end{equation}
where $\bm{\hat{F}_A}$, $\bm{\hat{F}_V} \in \mathbb{R}^{T \times D}$ are the updated audio and visual features, $\bm{W}^Q,\bm{W}^K,\bm{W}^V \in \mathbb{R}^{D \times D}$ are learnable parameters, $\delta$ is the softmax function.
Here, we simplify the presentation by omitting the feed-forward layers attached to the MHA operations.
The MHA mechanism allows each audio (visual) segment to engage with all the visual (audio) segments. 
The features of one modality can be augmented by incorporating relevant event information from the other modality through high cross-modal attention weights (\textit{event consistency}).

Furthermore, {we consider that each modality may contain event-unrelated background noise within specific temporal segments.
Each modality should focus on the key temporal regions of the counterpart modality that contain informative foreground.}
This is because a video segment contains an audio-visual event only if both the audio segment and visual segment include this event (\textit{temporal consistency}).
Therefore, we design \textbf{a Temporal Consistency-Gated branch (TCG)}.
Specifically, the feature of one modality is first enhanced by intra-modal attention, then it is processed through a {linear} layer followed by the sigmoid activation ($\sigma$).
This yields the temporal weight $\bm{g} \in \mathbb{R}^{T}$, which then serves as a consistency gate for the feature of the other modality.
These operations are formulated as, 
\begin{equation}\label{eq:gate}
\begin{gathered}
\bm{g}_V = \sigma((\text{MHA}(\bm{F}_V, \bm{F}_V, \bm{F}_V))\bm{W}_v), \\
\hat{\bm{F}}_A \Longleftarrow \hat{\bm{F}}_A + \bm{g}_V \odot  \hat{\bm{F}}_A, \\
\bm{g}_A = \sigma((\text{MHA}(\bm{F}_A, \bm{F}_A, \bm{F}_A))\bm{W}_a), \\
\hat{\bm{F}}_V \Longleftarrow \hat{\bm{F}}_V + \bm{g}_A \odot  \hat{\bm{F}}_V, \\
\end{gathered}
\end{equation}
where $\bm{W}_v, \bm{W}_a \in \mathbb{R}^{D \times 1}$ are learnable parameters of the linear layers, $\odot$ denotes element-wise multiplication, $\hat{\bm{F}}_A, \hat{\bm{F}}_V \in \mathbb{R}^{T \times D}$.
Note that the latent feature $\bm{g}$ is automatically learned during model training.
Under this guidance, the feature of each modality is further enhanced by focusing on the informative temporal regions recognized in the other modality.

For convenience, we denote the above operations of two branches in Eqs.~\ref{eq:MHA} and~\ref{eq:gate} as a Cross-Modal Consistency Collaboration (CMCC) layer, symbolized as,
\begin{equation}
\begin{gathered}
\bm{\hat{F}}_A, \bm{\hat{F}}_V = \text{CMCC}(\bm{F}_A, \bm{F}_V).
\end{gathered}
\end{equation}
Inspired by the baseline~\cite{geng2023dense}, we incorporate cross-modal collaboration at different temporal scales to better perceive events of varied durations.
Specifically, we utilize $L_c$ stacked CMCC layers.
Within each layer, the audio and visual features are first downsampled with a stride $2^{l_c-1}$, where $l_c$ is the index of the current layer.
The downsampled features are used as the \textit{query}, \textit{key}, and \textit{value} in the MHA operation (Eq.~\ref{eq:MHA}).
The output of the $l_c$-1 CMCC layer is then utilized as the input to the $l_c$ layer. Therefore, we obtain:
\begin{equation}
\begin{gathered}
\bm{\hat{F}}_A^{l_c}, \bm{\hat{F}}_V^{l_c} = \text{CMCC}(\bm{\hat{F}}_A^{l_c-1}, \bm{\hat{F}}_V^{l_c-1}),
\end{gathered}
\end{equation}
where $l_c \in \{1, 2, ..., L_c\}$, $\bm{\hat{F}}_A^0 = \bm{F}_A, \bm{\hat{F}}_V^0 = \bm{F}_V$, and $\bm{\hat{F}}_A^{l_c}, \bm{\hat{F}}_V^{l_c} \in \mathbb{R}^{T_{l_c} \times D}$ $(T_{l_c} = T / 2^{l_c-1})$.
Then, the audio and visual features at the same temporal scale are concatenated, resulting in the feature pyramid $\bm{Z} = \{\bm{Z}^{l_c}\}_{l_c=1}^{L_c}$, where $\bm{Z}^{l_c} = \text{Concat}(\bm{\hat{F}}_A^{l_c}, \bm{\hat{F}}_V^{l_c}) \in \mathbb{R}^{T_{l_c} \times 2D}$.

\subsection{Multi-Temporal Granularity Collaboration}
After obtaining the audiovisual features at different temporal scales, the previous baseline~\cite{geng2023dense} directly sends them into a Temporal Dependency Modelling (TDM) module, which models the dependencies of simultaneous events and consecutive segments.
However, the TDM operates on features at {separate} temporal scales.
In contrast, we propose {a very simple but effective} Multi-Temporal Granularity Collaboration (MTGC) module to enhance the collaboration among different temporal scales.
Our motivation is that the \textit{coarse} temporal features ($\bm{Z}^{l_c}$ with larger $l_c$) have a larger receptive field for recognizing the occurring event, while the \textit{fine}-grained features ($\bm{Z}^{l_c}$ with smaller $l_c$) are more beneficial for precise event boundary prediction.
Our proposed MTGC module explores a bidirectional collaboration mechanism for better temporal localization, as detailed next.

\noindent\textbf{Coarse-to-Fine Collaboration (C2F).}
Given the concatenated audiovisual features $\bm{Z}^{k} \in \mathbb{R}^{T_k \times 2D}$ ($T_k=T/2^{k-1}$) at the $k$-th temporal granularity ($k \in \{1, 2,..., L_c-1\}$), we treat it as the current \textit{fine}-grained temporal feature.
Then, the feature at adjacent $k+1$ granularity $\bm{Z}^{k+1}$ can be regarded as the \textit{coarse}-grained feature.
We then apply a linear layer followed by the {ReLU} activation to transform the coarse-grained feature to match the dimension of the fine-grained feature $\bm{Z}^{k}$, written as,
\begin{equation}
\begin{split}
    \bm{U}^k = \text{ReLU}(\bm{W}_u\bm{Z}^{k+1}),
\end{split}
\end{equation}
where $\bm{W}_u \in \mathbb{R}^{T_k \times T_{k+1}}$ is the learnable paramter of the linear layer, $\bm{U}^k \in \mathbb{R}^{T_k \times 2D}$.
In principle, $\bm{U}^k$ provides the event information from a more coarse-grained level, which can collaborate with the fine-grained $\bm{Z}^k$.
We achieve the coarse-to-fine collaboration between $\bm{U}^k$ and $\bm{Z}^k$ via simple multi-head attention (MHA), thus generating the updated feature at $k$-th temporal granularity $\bm{G}^k$: 
\begin{equation}
\begin{split}\bm{G}^k = \text{MHA}(\bm{U}^k, \bm{Z}^k, \bm{Z}^k),
\end{split}
\end{equation}
where $\bm{G}^k \in \mathbb{R}^{T_k \times D}, k=\{1,2,...,L_c-1\}$.
Notably, for the largest $L_c$-th granularity, there are no more coarse-grained temporal features, so we set $\bm{G}^{L_c} = \bm{Z}^{L_c}$.
After the C2F collaboration among temporal features at different granularities, we adopt the TDM~\cite{geng2023dense} module to further enhance features at each separate temporal scale.

\noindent\textbf{Fine-to-Coarse Collaboration (F2C).}
In addition to the coarse-to-fine collaboration, we also develop a fine-to-coarse collaboration mechanism that enables a bidirectional interaction for features at multiple temporal granularities.
Assuming the updated feature $\bm{G}^m \in \mathbb{R}^{T/2^{m-1} \times 2D}$ at the $m$-th temporal granularity ($m \in \{ 2,..., L_c\}$) as \textit{coarse}-grained feature, the feature $\bm{G}^{m-1} \in \mathbb{R}^{T/2^{m-2} \times 2D}$ at the adjacent $m-1$ temporal granularity can be regarded as the \textit{fine}-grained feature.

We first temporally downsample the fine-grained feature $\bm{G}^{m-1}$ via max-pooling to align its dimension with $\bm{G}^{m}$. 
Then, we model the fine-to-coarse collaboration using multi-head attention.
These operations can be summarized as,
\begin{equation}
\begin{gathered}
 \bm{\hat{G}}^{m-1} = \text{MaxPooling}(\bm{G}^{m-1}), \\
    \bm{O}^m = \bm{G}^m + \text{MHA}( \bm{G}^m, \bm{\hat{G}}^{m-1}, \bm{\hat{G}}^{m-1}),
\end{gathered}
\end{equation}
where $\bm{O}^m \in \mathbb{R}^{T/2^{m-1} \times 2D}$ is the updated feature at the $m$-th temporal granularity, $m=2,...,L_c$.
Let $\bm{O}^m = \text{F2C}(\bm{G}^{m-1}, \bm{G}^m)$ represent the fine-to-coarse collaboration process described above, the output $\bm{O}^m$ is used as the \textit{fine}-grained feature in the next $m$+1 granularity:
$\bm{O}^{m+1} = \text{F2C}(\bm{O}^m, \bm{G}^{m+1})$.
For the temporal granularity $m=1$, there are no more fine-grained features, we simply assign $\bm{O}^{1}=\bm{G}^1$.
In this way, the temporal features at each granularity $\bm{O}^{l_c} (l_c=1,2,...,L_c)$ are enhanced by incorporating both coarse- and fine-grained event clues, which are ready for decoding audio-visual events across varied temporal ranges.

\subsection{Decoder}
Following the paradigm of baseline~\cite{geng2023dense}, the decoder of the DAVEL task includes a classification head and a regression head.
Given the feature $\bm{O}^{l_c} \in \mathbb{R}^{T/2^{l_c-1} \times 2D}$ at the $l_c$ temporal granularity ($l_c=\{1,2,..., L_c\}$), the classification head predicts the corresponding event probability $p(c_t)$ for each timestamp $t$.
The classification head is implemented by three 1D convolution layers following a sigmoid function.
As for the regression head, it also consists of three 1D convolutions but is activated with the ReLU function.
This head directly regresses the distances from the current timestamp $t$ to the start and end timestamp of an event $(d_{s,t}, d_{e,t})$ if the event exists.
The regression output with the shape of $[2, C, T_{l_c}]$ indicates the onsets and offsets to an event at each timestamp, which is also class-aware for recognizing overlapping events with different categories.

\noindent\textbf{Training.} 
We train our model by employing two losses, \ie, a focal loss~\cite{Lin2017FocalLF} $\mathcal{L}_{cls}$ for imbalanced event classification, and a generalized IoU loss~\cite{Rezatofighi2019GeneralizedIO} $\mathcal{L}_{reg}$ for distance regression.
The total training objective can be written as, 
\begin{equation}
\begin{gathered}
\mathcal{L} = \alpha \sum_{t} \mathcal{L}_{cls}+ \beta \sum_{t}\mathds{I}_t\mathcal{L}_{reg},
\end{gathered}
\end{equation}
where $\alpha$ and $\beta$ are two hyperparameters, which are identical to those in baseline~\cite{geng2023dense}, $\mathds{I}_t$ is a function indicating whether a timestamp $t$ contains audio-visual events.

\noindent\textbf{Inference.}
For each timestamp, we predict its event classes and the temporal boundary of each event following Eq.~\ref{eq:problem_statement}. 
The results are then post-processed using the Soft-NMS~\cite{Bodla2017SoftNMSI} technique to suppress predictions that are predicted to be in the same category but are highly overlapping.

\section{Experiments}
\subsection{Experimental Setups}
\textbf{Dataset.}
Our experiments are conducted on the official UnAV-100~\cite{geng2023dense} dataset, specifically constructed for the dense audio-visual event localization task.
1) \textit{Untrimmed videos.} The UnAV-100 dataset comprises 10,790 untrimmed videos with varied temporal lengths, with the majority exceeding 40 seconds. 
2) \textit{Multiple categories. }The videos encompass 100 categories of audio-visual events commonly found in natural environments, including human or animal activities, musical instruments, various vehicles, \etc.
\textit{3) Densely overlapping events of varying durations.}
On average, each video features 2.8 audio-visual events, highlighting the presence of multiple overlapping events. 
Furthermore, these events span a range of durations (\ie, distinct temporal windows). 
These characteristics make event localization challenging.
Following the standard dataset split, the distribution among training, validation, and testing subsets is set at a ratio of 3:1:1.

\noindent\textbf{Evaluation metric.}
Following the baseline~\cite{geng2023dense}, we adopt the mean Average Precision (mAP) as the metric for evaluating temporal localization results.
We report mAPs at tIoU thresholds ranging from 0.5 to 0.9 in increments of 0.1 ([0.5:0.1:0.9]).
Additionally, we report the average mAP (denoted as {`Avg.'}), calculated across an expanded range of thresholds [0.1:0.1:0.9], serving as a comprehensive measure for comparing overall model performance.

\noindent\textbf{Implementation details.}
For visual feature extraction, frames are sampled at 25 FPS for each video, and the RAFT~\cite{teed2020raft} is utilized to extract the optical flow.
Then, 24 consecutive RGB and the optimal flow frames are sent into the two-stream pretrained I3D~\cite{carreira2017quo} model, yielding 2048-D visual features.
For audio feature extraction, audio signals are first split every 0.96s using a sliding window of 0.32s. Then, the pretrained VGGish~\cite{hershey2017vggish} is used to extract features for each audio segment, resulting in the 128-D audio features.
It is noteworthy that the video sequences vary in length; thus, we crop or pad the extracted audio and visual features to a maximum length of $T$=224.
The layer numbers $L_s$ and $L_c$ are empirically set to 2 and 6, respectively.
We train our model for 40 epochs with a batch size of 16.
The Adam optimizer is used, with the initial learning rate and the weight decay set to 1e-4.
Experiments are conducted on a NVIDIA A40 GPU.

\begin{table}[t]
  \setlength{\tabcolsep}{1mm}
      \resizebox{\linewidth}{!}{
  \begin{tabular}{cccccccc}
    \toprule
    Methods   &0.5 & 0.6 &0.7 &0.8 &0.9 &Avg.\\ \midrule
    VSGN~\cite{zhao2021video}  & 24.5 & 20.2 &15.9 &11.4 &6.8 &24.1\\
    TadTR~\cite{liu2022end}  & 30.4 &27.1 &23.3 &19.4 &14.3 &29.4\\
    ActionFormer~\cite{zhang2022actionformer}  &43.5 &39.4 &33.4 &27.3 &17.9 &42.2\\ 
    DAVEL~\cite{geng2023dense} & 50.6 &45.8 &39.8 &32.4 &21.1 &47.8\\
    \textbf{CCNet (ours)} & \textbf{51.9} & \textbf{47.2} & \textbf{41.5} & \textbf{34.1} & \textbf{23.0} & \textbf{49.2} \\ \midrule
    $^\vartriangle$DAVEL~\cite{geng2023dense} & 53.8 & 48.7 & 42.2 & 33.8 & 20.4 & 51.0 \\
    $^\vartriangle$UniAV~\cite{geng2024uniav}  &54.8 &49.4 &43.2 &35.3 &22.5 &51.7 \\
    $^\vartriangle$\textbf{CCNet (ours)} & \textbf{57.3} & \textbf{52.2} & \textbf{46.2} & \textbf{38.1} & \textbf{25.6} & \textbf{54.1} \\ 
  \bottomrule
\end{tabular}
}
  \caption{\textbf{Comparison with prior works.} `$^\vartriangle$' denotes that more advanced audio and visual features extracted by  ONE-PEACE~\cite{wang2023one} are used.
  }
  \label{tab:sota_comparsion}
\end{table}

\begin{figure}[t]
    \centering
    \includegraphics[width=0.92\linewidth]{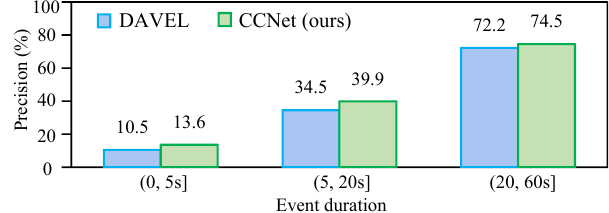}
    \caption{\textbf{Localization results for various durations.}
    }
\label{fig:different_temopral_scales}
\end{figure}

\subsection{Comparison to Prior Works}
To ensure a more comprehensive comparison, we also compare our method with other representative approaches tailored for the temporal localization task, namely VSGN~\cite{zhao2021video}, TadTR~\cite{liu2022end}, and ActionFormer~\cite{zhang2022actionformer}.
Given that we focus on a multimodal task, these methods are adapted to utilize concatenated audio and visual features as inputs.
As shown in Table~\ref{tab:sota_comparsion}, these temporal action localization models significantly lag behind the baseline DAVEL~\cite{geng2023dense}.
This underscores the critical importance of specialized designs for cross-modal relation modeling in the dense audio-visual event localization task.
Furthermore, our method is superior to the baseline DAVEL~\cite{geng2023dense} method.
We attain a new state-of-the-art performance, achieving an average mAP of 49.2\%.
Our method surpasses the baseline DAVEL across all tIoUs thresholds, with notable improvements of 1.9\% at the stringent tIoU=0.9.
Furthermore, our method can be significantly enhanced by adopting more advanced audio-visual features extracted using ONE-PEACE~\cite{wang2023one}, continuing to surpass previous baselines.
These results demonstrate the superiority of our method, attributed to the proposed two core modules that facilitate cross-modal consistency and multi-temporal granularity collaborations.

In addition, we compare the proposed method with the baseline DAVEL~\cite{geng2023dense} on the localization of events in different temporal durations.
We analyze the events in videos from the UnAV-100 dataset, focusing on the event durations at $(0s, 5s]$ (short), $(5s, 20s]$ (middle), and $(20s, 60s]$ (long).
For one event in a specific duration, we consider this event correctly localized if the tIoU between the model prediction and the ground truth exceeds a threshold of 0.5 and the predicted event category is correct.
Then, we calculate the percentage of the correct localized events relative to the total number of events within that duration (\textit{precision}).
As shown in Fig.~\ref{fig:different_temopral_scales}, our method surpasses the baseline in event localization across varied temporal durations.
Particularly, our method achieves a 5.4\% improvement over the baseline for events with a duration of $(5s, 20s]$.
These improvements can be attributed to the multi-temporal granularity collaboration module in our method, which enables the effective integration of temporal cues across various scales.

\begin{table}[t]
  
\small
\centering
\setlength{\tabcolsep}{2mm}
      \resizebox{\linewidth}{!}{
  \begin{tabular}{ccccccccc}
    \toprule
    \multirow{2}{*}{CMCC} & \multicolumn{2}{c}{MTGC} & \multirow{2}{*}{0.5} & \multirow{2}{*}{0.6} & \multirow{2}{*}{0.7} &\multirow{2}{*}{0.8} &\multirow{2}{*}{0.9} &\multirow{2}{*}{Avg.}\\
    \cmidrule{2-3}
    & C2F & F2C & & & & & & \\ \midrule
    \ding{52} & \ding{56} & \ding{56} & 50.5 & 45.6 & 39.8 & 33.1 & 22.9 & 47.9 \\
    \ding{52} & \ding{52} & \ding{56}  & 51.3 & 46.1 & 40.0 & 32.8 & 22.4 & 48.3 \\
    \ding{52} & \ding{56} & \ding{52} & 50.8 & 46.2 & 40.6 & 34.0 & \textbf{23.4} & 48.2 \\
    \ding{52} & \ding{52} & \ding{52} & \textbf{51.9} & \textbf{47.2} & \textbf{41.5} & \textbf{34.1} & 23.0 & \textbf{49.2}\\
  \bottomrule
\end{tabular}}
\caption{\textbf{The ablation study of our core modules.}
}
\label{tab:ablation_core_modules}
\end{table}

\begin{figure*}[t]
  \centering
\includegraphics[width=\linewidth]{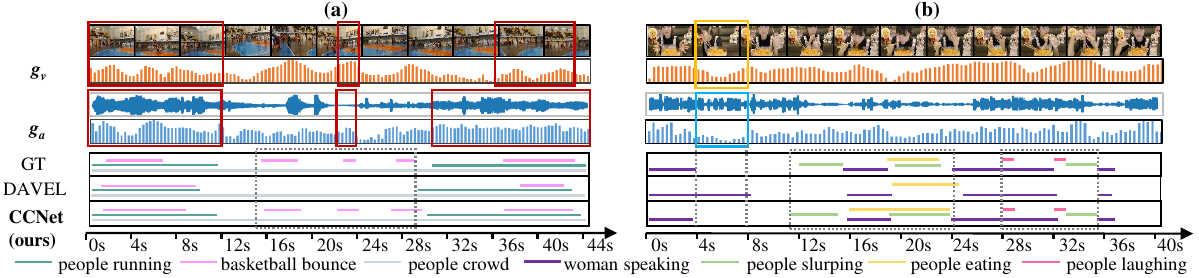}
  \caption{\textbf{Qualitative examples of dense audio-visual event localization.}}
  \label{fig:davel_examples}
\end{figure*}

\subsection{Ablation Studies}
\textbf{Effectiveness of our core modules.}
{Our method consists of the core Cross-Modal Consistency Collaboration (CMCC) and Multi-Temporal Granularity Collaboration (MTGC) modules.
Specifically, the MTGC module encompasses the C2F and F2C collaboration blocks.
We conduct ablation experiments to explore their impacts.
As presented in Table~\ref{tab:ablation_core_modules}, the average mAP is 47.9\% when utilizing only the CMCC module. Note that this variant model still marginally outperforms the baseline DAVEL model regarding the average mAP. Moreover, this variant model exceeds DAVEL by 1.8\% in precision at the tIoU threshold of 0.9, highlighting our superior localization capabilities.
While DAVEL also considers encoding cross-modal relations, our CMCC module additionally incorporates temporal consistency within audio and visual modalities.
The model performance can be improved by separately adding the C2F or the F2C block, indicating the effectiveness of each collaboration mechanism.
Ultimately, our model achieves the highest performance when employing both C2F and F2C blocks simultaneously.
This suggests that the bidirectional C2F and F2C work collaboratively, contributing to better exploitation from multi-temporal granularities.}

\noindent\textbf{Ablation study of the CMCC module.} 
We assess the effectiveness of each branch of the CMCC module: the Cross-Modal Interaction (CMI) branch and the Temporal Consistency-Gated (TCG) branch.
The experimental results are shown in the lower part of Table~\ref{tab:ablation_CMCC}.
We find that each branch is beneficial for improving event localization performance.
The CMI branch facilitates one modality in aggregating relevant or complementary event semantics from the other modality;
while the TCG branch enables one modality to recognize the key temporal regions in the other modality. 
Consequently, the model performance can be enhanced by utilizing these two branches simultaneously.

\begin{table}[t]
\small
  \setlength{\tabcolsep}{3mm}
      \resizebox{\linewidth}{!}{
  \begin{tabular}{cccccccc}
    \toprule
  CMI & TCG & 0.5 & 0.6 & 0.7 &0.8 &0.9 &Avg.\\
  \midrule
    \ding{52} & \ding{52}  & \textbf{51.9} & \textbf{47.2} & \textbf{41.5} & \textbf{34.1} & \textbf{23.0} & \textbf{49.2}\\
    \ding{52} & \ding{56}  & 50.8 & 46.1 & 40.6 & 33.6 & 22.6 & 48.1\\
    \ding{56} & \ding{52}  & 49.9 & 45.3 & 40.0 & 33.2 & 22.8 & 47.6\\
    \bottomrule
\end{tabular}}
   \caption{\textbf{The ablation results of our CMCC module. } 
  We ablate the two branches in CMCC: the abbreviation `CMI' is short for the Cross-Modal Interaction branch, while `TCG' represents the Temporal Consistency-Gated branch.}
  \label{tab:ablation_CMCC}
\end{table}

\noindent\textbf{Ablation study of the MTGC module.}
In the proposed MTGC module, the Coarse-to-Fine (C2F) block is implemented first, followed by the Fine-to-Coarse (F2C) collaboration block.
Here, we explore the impacts of the operational sequence of these two blocks.
The results, as shown in Table~\ref{tab:c2f_f2c_order}, indicate that applying C2F before F2C results in a higher average mAP of 49.2\%, whereas the reversed order leads to a lower performance of 47.1\%.
This suggests that the order in which the C2F and F2C blocks are applied influences the overall performance of the MTGC module. In our supplementary material, we provide additional ablation studies on the proposed CMCC and MTGC modules.

\begin{table}[t]
\small
  \setlength{\tabcolsep}{3mm}
      \resizebox{\linewidth}{!}{
  \begin{tabular}{ccccccc}
    \toprule
    Strategies &0.5 & 0.6 &0.7 &0.8 &0.9 &Avg.\\
    \midrule
    C2F $\rightarrow$ F2C & \textbf{51.9} & \textbf{47.2} & \textbf{41.5} & \textbf{34.1} & \textbf{23.0} & \textbf{49.2}\\
    F2C $\rightarrow$ C2F & 49.9 & 45.3 & 40.1 & 32.7 & 21.8 & 47.1\\
  \bottomrule
\end{tabular}}
\caption{\textbf{Ablation study on the operational sequence of C2F and F2C collaboration blocks in MTGC module.} `A' $\rightarrow$ `B' denotes the initial application of block `A', followed by the execution of block `B' in MTGC module implementation.}
\label{tab:c2f_f2c_order}
\end{table}

\subsection{Qualitative Results}
We present some qualitative examples of dense audio-visual event localization.
Fig.~\ref{fig:davel_examples} (a) illustrates a video sample containing three classes of audio-visual events: \textit{people running}, \textit{basketball bounce}, and \textit{people crowd}, densely distributed across various temporal extents.
Compared to the baseline model DAVEL~\cite{geng2023dense}, our method demonstrates superior performance in temporally localizing events with varying durations.
For instance, DAVEL fails to recognize event \textit{basketball bounce} within the 15s$\sim$30s (marked by the gray dotted box). In contrast, our method not only successfully identifies this event but also provides satisfactory temporal boundaries, highlighting the advantages of our proposed {multi-temporal granularity collaboration} mechanism.
We also plot the curves of the learned temporal consistency gates $\bm{g}_A$ and $\bm{g}_V$, which indeed assign higher weights to temporal regions associated with these events (highlighted by the red boxes in the figure).
In Fig.~\ref{fig:davel_examples}(b), the baseline DAVEL overlooks the audio-visual events \textit{people slurping} and \textit{people laughing} (gray dotted box).
Conversely, our method accurately predicts the event categories and determines precise start and end timestamps. 
Furthermore, DAVEL incorrectly identifies an audio-visual event, \textit{woman speaking}, within the 4s$\sim$8s.
However, our model's temporal consistency gate $\bm{g}_A$ assigns very low weights to this period (blue box), indicating the absence of audio events.
Despite the clear depiction of \textit{woman speaking} in the visual frames, as indicated by the high $\bm{g}_V$ values during this period (yellow box), the lack of corresponding audio events evidenced by our model ensures accurate prediction for this audio-visual event.
These results confirm the effectiveness of the proposed {cross-modal consistency collaboration} mechanism.

\section{Conclusion}
We tackle a practical task of dense audio-visual event localization, which aims to temporally localize the audio-visual events densely occurring in untrimmed audible videos.
We introduce a new CCNet approach and formulate its core modules with consideration of two essential perspectives: Cross-Modal Consistency Collaboration (CMCC) and Multi-Temporal Granularity Collaboration (MTGC).
The CMCC module utilizes a cross-modal interaction branch to encode audio-visual interactions and incorporates a temporal consistency-gated branch to regulate each modality's focus on event-related temporal regions.
The MTGC module consists of a coarse-to-fine and a fine-to-coarse collaboration block, which are beneficial for bidirectional cooperation among coarse- and fine-grained features across various temporal granularities.
Experimental results demonstrate that our method surpasses previous baselines in accurately localizing dense audio-visual events of varying durations.

\section*{Acknowledgements }
We would like to express our sincere gratitude to the anonymous reviewers for their invaluable comments and insightful suggestions.
The computation is completed on the HPC Platform of Hefei University of Technology.
This work was supported by the National Natural Science Foundation of China (62272144), the Major Project of Anhui Province (2408085J040), and the Fundamental Research Funds for the Central Universities (JZ2024HGTG0309, JZ2024AHST0337).

\bibliography{aaai25}

\begin{thebibliography}{54}
\providecommand{\natexlab}[1]{#1}

\bibitem[{Bodla et~al.(2017)Bodla, Singh, Chellappa, and Davis}]{Bodla2017SoftNMSI}
Bodla, N.; Singh, B.; Chellappa, R.; and Davis, L.~S. 2017.
\newblock Soft-NMS — Improving Object Detection with One Line of Code.
\newblock In \emph{ICCV}, 5562--5570.

\bibitem[{Carreira and Zisserman(2017)}]{carreira2017quo}
Carreira, J.; and Zisserman, A. 2017.
\newblock Quo vadis, action recognition? a new model and the kinetics dataset.
\newblock In \emph{CVPR}, 6299--6308.

\bibitem[{Cheng et~al.(2022)Cheng, Liu, Zhou, Qian, Wu, and Wang}]{cheng2022joint}
Cheng, H.; Liu, Z.; Zhou, H.; Qian, C.; Wu, W.; and Wang, L. 2022.
\newblock Joint-Modal Label Denoising for Weakly-Supervised Audio-Visual Video Parsing.
\newblock In \emph{ECCV}, 431--448.

\bibitem[{Gao, Chen, and Xu(2023)}]{gao2023collecting}
Gao, J.; Chen, M.; and Xu, C. 2023.
\newblock Collecting Cross-Modal Presence-Absence Evidence for Weakly-Supervised Audio-Visual Event Perception.
\newblock In \emph{CVPR}, 18827--18836.

\bibitem[{Geng et~al.(2023)Geng, Wang, Duan, Cong, and Zheng}]{geng2023dense}
Geng, T.; Wang, T.; Duan, J.; Cong, R.; and Zheng, F. 2023.
\newblock Dense-localizing audio-visual events in untrimmed videos: A large-scale benchmark and baseline.
\newblock In \emph{CVPR}, 22942--22951.

\bibitem[{Geng et~al.(2024)Geng, Wang, Zhang, Duan, Guan, and Zheng}]{geng2024uniav}
Geng, T.; Wang, T.; Zhang, Y.; Duan, J.; Guan, W.; and Zheng, F. 2024.
\newblock UniAV: Unified Audio-Visual Perception for Multi-Task Video Localization.
\newblock \emph{arXiv preprint arXiv:2404.03179}.

\bibitem[{Guo et~al.(2023)Guo, Ying, Chen, Niu, Li, Qu, Qi, Zhou, Xing, Yue, Shi, Wang, Zhang, and Liang}]{guo2023audio}
Guo, R.; Ying, X.; Chen, Y.; Niu, D.; Li, G.; Qu, L.; Qi, Y.; Zhou, J.; Xing, B.; Yue, W.; Shi, J.; Wang, Q.; Zhang, P.; and Liang, B. 2023.
\newblock Audio-Visual Instance Segmentation.
\newblock \emph{arXiv preprint arXiv:2310.18709}.

\bibitem[{Heilbron et~al.(2015)Heilbron, Escorcia, Ghanem, and Niebles}]{Heilbron2015ActivityNetAL}
Heilbron, F.~C.; Escorcia, V.; Ghanem, B.; and Niebles, J.~C. 2015.
\newblock ActivityNet: A large-scale video benchmark for human activity understanding.
\newblock \emph{CVPR}, 961--970.

\bibitem[{Hershey et~al.(2017)Hershey, Chaudhuri, Ellis, Gemmeke, Jansen, Moore, Plakal, Platt, Saurous, Seybold et~al.}]{hershey2017vggish}
Hershey, S.; Chaudhuri, S.; Ellis, D.~P.; Gemmeke, J.~F.; Jansen, A.; Moore, R.~C.; Plakal, M.; Platt, D.; Saurous, R.~A.; Seybold, B.; et~al. 2017.
\newblock CNN architectures for large-scale audio classification.
\newblock In \emph{ICASSP}, 131--135.

\bibitem[{Hou et~al.(2024)Hou, Li, Tian, and Hu}]{10.1145/3672079}
Hou, W.; Li, G.; Tian, Y.; and Hu, D. 2024.
\newblock Towards Long Form Audio-visual Video Understanding.
\newblock \emph{ACM Trans. Multimedia Comput. Commun. Appl.}, 1--22.

\bibitem[{Hu, Nie, and Li(2019)}]{hu2019deep}
Hu, D.; Nie, F.; and Li, X. 2019.
\newblock Deep multimodal clustering for unsupervised audiovisual learning.
\newblock In \emph{CVPR}, 9248--9257.

\bibitem[{Iashin and Rahtu(2020)}]{iashin2020better}
Iashin, V.; and Rahtu, E. 2020.
\newblock A better use of audio-visual cues: Dense video captioning with bi-modal transformer.
\newblock \emph{arXiv preprint arXiv:2005.08271}.

\bibitem[{Jiang et~al.(2022)Jiang, Xu, Chen, Zhang, Song, Shen, Lu, and Shen}]{jiang2022dhhn}
Jiang, X.; Xu, X.; Chen, Z.; Zhang, J.; Song, J.; Shen, F.; Lu, H.; and Shen, H.~T. 2022.
\newblock DHHN: Dual Hierarchical Hybrid Network for Weakly-Supervised Audio-Visual Video Parsing.
\newblock In \emph{ACM MM}, 719--727.

\bibitem[{Lao et~al.(2023)Lao, Pu, Liu, He, Bakker, and Lew}]{lao2023coca}
Lao, M.; Pu, N.; Liu, Y.; He, K.; Bakker, E.~M.; and Lew, M.~S. 2023.
\newblock COCA: COllaborative CAusal Regularization for Audio-Visual Question Answering.
\newblock In \emph{AAAI}, 12995--13003.

\bibitem[{Li, Hou, and Hu(2023)}]{li2023progressive}
Li, G.; Hou, W.; and Hu, D. 2023.
\newblock Progressive Spatio-temporal Perception for Audio-Visual Question Answering.
\newblock In \emph{ACM MM}, 7808--7816.

\bibitem[{Li et~al.(2022)Li, Wei, Tian, Xu, Wen, and Hu}]{li2022learning}
Li, G.; Wei, Y.; Tian, Y.; Xu, C.; Wen, J.-R.; and Hu, D. 2022.
\newblock Learning to Answer Questions in Dynamic Audio-Visual Scenarios.
\newblock In \emph{CVPR}, 19108--19118.

\bibitem[{Li et~al.(2023)Li, Yang, Chen, Yang, and Xiao}]{li2023catr}
Li, K.; Yang, Z.; Chen, L.; Yang, Y.; and Xiao, J. 2023.
\newblock Catr: Combinatorial-dependence audio-queried transformer for audio-visual video segmentation.
\newblock In \emph{ACM MM}, 1485--1494.

\bibitem[{Li et~al.(2024{\natexlab{a}})Li, Guo, Zhou, Zhang, and Wang}]{li2024object}
Li, Z.; Guo, D.; Zhou, J.; Zhang, J.; and Wang, M. 2024{\natexlab{a}}.
\newblock Object-Aware Adaptive-Positivity Learning for Audio-Visual Question Answering.
\newblock In \emph{AAAI}, 3306--3314.

\bibitem[{Li et~al.(2024{\natexlab{b}})Li, Zhou, Zhang, Tang, Li, and Guo}]{li2024post}
Li, Z.; Zhou, J.; Zhang, J.; Tang, S.; Li, K.; and Guo, D. 2024{\natexlab{b}}.
\newblock Patch-level Sounding Object Tracking for Audio-Visual Question Answering.
\newblock \emph{arXiv preprint arXiv:2412.10749}.

\bibitem[{Lin et~al.(2017)Lin, Goyal, Girshick, He, and Doll{\'a}r}]{Lin2017FocalLF}
Lin, T.-Y.; Goyal, P.; Girshick, R.~B.; He, K.; and Doll{\'a}r, P. 2017.
\newblock Focal Loss for Dense Object Detection.
\newblock In \emph{ICCV}, 2999--3007.

\bibitem[{Lin et~al.(2021)Lin, Tseng, Lee, Lin, and Yang}]{lin2021exploring}
Lin, Y.-B.; Tseng, H.-Y.; Lee, H.-Y.; Lin, Y.-Y.; and Yang, M.-H. 2021.
\newblock Exploring Cross-Video and Cross-Modality Signals for Weakly-Supervised Audio-Visual Video Parsing.
\newblock In \emph{NeurIPS}.

\bibitem[{Liu et~al.(2023)Liu, Li, Qi, Zhang, Li, Wang, and Yu}]{liu2023audio}
Liu, C.; Li, P.~P.; Qi, X.; Zhang, H.; Li, L.; Wang, D.; and Yu, X. 2023.
\newblock Audio-Visual Segmentation by Exploring Cross-Modal Mutual Semantics.
\newblock In \emph{ACM MM}, 7590--7598.

\bibitem[{Liu et~al.(2022)Liu, Wang, Hu, Tang, Zhang, Bai, and Bai}]{liu2022end}
Liu, X.; Wang, Q.; Hu, Y.; Tang, X.; Zhang, S.; Bai, S.; and Bai, X. 2022.
\newblock End-to-end temporal action detection with transformer.
\newblock \emph{TIP}, 31: 5427--5441.

\bibitem[{Mao et~al.(2024)Mao, Shen, Zhang, Qin, Zhou, Xiang, Zhong, and Dai}]{mao2024tavgbench}
Mao, Y.; Shen, X.; Zhang, J.; Qin, Z.; Zhou, J.; Xiang, M.; Zhong, Y.; and Dai, Y. 2024.
\newblock TAVGBench: Benchmarking text to audible-video generation.
\newblock In \emph{ACM MM}, 6607--6616.

\bibitem[{Mao et~al.(2023)Mao, Zhang, Xiang, Zhong, and Dai}]{mao2023multimodal}
Mao, Y.; Zhang, J.; Xiang, M.; Zhong, Y.; and Dai, Y. 2023.
\newblock Multimodal Variational Auto-encoder based Audio-Visual Segmentation.
\newblock In \emph{ICCV}, 954--965.

\bibitem[{Qian et~al.(2020)Qian, Hu, Dinkel, Wu, Xu, and Lin}]{qian2020multiple}
Qian, R.; Hu, D.; Dinkel, H.; Wu, M.; Xu, N.; and Lin, W. 2020.
\newblock Multiple Sound Sources Localization from Coarse to Fine.
\newblock In \emph{ECCV}, 1--16.

\bibitem[{Rezatofighi et~al.(2019)Rezatofighi, Tsoi, Gwak, Sadeghian, Reid, and Savarese}]{Rezatofighi2019GeneralizedIO}
Rezatofighi, S.~H.; Tsoi, N.; Gwak, J.; Sadeghian, A.; Reid, I.~D.; and Savarese, S. 2019.
\newblock Generalized Intersection Over Union: A Metric and a Loss for Bounding Box Regression.
\newblock In \emph{CVPR}, 658--666.

\bibitem[{Senocak et~al.(2021)Senocak, Oh, Kim, Yang, and Kweon}]{senocak2021TAPMI}
Senocak, A.; Oh, T.-H.; Kim, J.; Yang, M.-H.; and Kweon, I.~S. 2021.
\newblock Learning to Localize Sound Sources in Visual Scenes: Analysis and Applications.
\newblock In \emph{TPAMI}, 1605--1619.

\bibitem[{Shen et~al.(2023)Shen, Li, Zhou, Qin, He, Han, Li, Dai, Kong, Wang et~al.}]{shen2023fine}
Shen, X.; Li, D.; Zhou, J.; Qin, Z.; He, B.; Han, X.; Li, A.; Dai, Y.; Kong, L.; Wang, M.; et~al. 2023.
\newblock Fine-grained audible video description.
\newblock In \emph{CVPR}, 10585--10596.

\bibitem[{Shi et~al.(2023)Shi, Zhong, Cao, Ma, Li, and Tao}]{Shi_2023_CVPR}
Shi, D.; Zhong, Y.; Cao, Q.; Ma, L.; Li, J.; and Tao, D. 2023.
\newblock TriDet: Temporal Action Detection With Relative Boundary Modeling.
\newblock In \emph{CVPR}, 18857--18866.

\bibitem[{Teed and Deng(2020)}]{teed2020raft}
Teed, Z.; and Deng, J. 2020.
\newblock Raft: Recurrent all-pairs field transforms for optical flow.
\newblock In \emph{ECCV}, 402--419.

\bibitem[{Tian et~al.(2018{\natexlab{a}})Tian, Guan, Goodman, Moore, and Xu}]{tian2018attempt}
Tian, Y.; Guan, C.; Goodman, J.; Moore, M.; and Xu, C. 2018{\natexlab{a}}.
\newblock An attempt towards interpretable audio-visual video captioning.
\newblock \emph{arXiv preprint arXiv:1812.02872}.

\bibitem[{Tian, Li, and Xu(2020)}]{tian2020unified}
Tian, Y.; Li, D.; and Xu, C. 2020.
\newblock Unified multisensory perception: Weakly-supervised audio-visual video parsing.
\newblock In \emph{ECCV}, 436--454.

\bibitem[{Tian et~al.(2018{\natexlab{b}})Tian, Shi, Li, Duan, and Xu}]{tian2018audio}
Tian, Y.; Shi, J.; Li, B.; Duan, Z.; and Xu, C. 2018{\natexlab{b}}.
\newblock Audio-visual event localization in unconstrained videos.
\newblock In \emph{ECCV}, 247--263.

\bibitem[{Tian et~al.(2018{\natexlab{c}})Tian, Shi, Li, Duan, and Xu}]{Tian_2018_ECCV}
Tian, Y.; Shi, J.; Li, B.; Duan, Z.; and Xu, C. 2018{\natexlab{c}}.
\newblock Audio-Visual Event Localization in Unconstrained Videos.
\newblock In \emph{ECCV}, 1--17.

\bibitem[{Vaswani et~al.(2017)Vaswani, Shazeer, Parmar, Uszkoreit, Jones, Gomez, Kaiser, and Polosukhin}]{vaswani2017attention}
Vaswani, A.; Shazeer, N.; Parmar, N.; Uszkoreit, J.; Jones, L.; Gomez, A.~N.; Kaiser, {\L}.; and Polosukhin, I. 2017.
\newblock Attention is all you need.
\newblock In \emph{NeurIPS}, 1--15.

\bibitem[{Wang et~al.(2023)Wang, Wang, Lin, Bai, Zhou, Zhou, Wang, and Zhou}]{wang2023one}
Wang, P.; Wang, S.; Lin, J.; Bai, S.; Zhou, X.; Zhou, J.; Wang, X.; and Zhou, C. 2023.
\newblock ONE-PEACE: Exploring One General Representation Model Toward Unlimited Modalities.
\newblock \emph{arXiv preprint arXiv:2305.11172}.

\bibitem[{Wu and Yang(2021)}]{wu2021exploring}
Wu, Y.; and Yang, Y. 2021.
\newblock Exploring heterogeneous clues for weakly-supervised audio-visual video parsing.
\newblock In \emph{CVPR}, 1326--1335.

\bibitem[{Xia and Zhao(2022)}]{9880210}
Xia, Y.; and Zhao, Z. 2022.
\newblock Cross-modal Background Suppression for Audio-Visual Event Localization.
\newblock In \emph{CVPR}, 19957--19966.

\bibitem[{Yang et~al.(2022)Yang, Wang, Duan, Chen, Hou, Jin, and Zhu}]{yang2022avqa}
Yang, P.; Wang, X.; Duan, X.; Chen, H.; Hou, R.; Jin, C.; and Zhu, W. 2022.
\newblock Avqa: A dataset for audio-visual question answering on videos.
\newblock In \emph{ACM MM}, 3480--3491.

\bibitem[{Yu et~al.(2022{\natexlab{a}})Yu, Cheng, Zhao, Feng, and Zhang}]{yu2021mm}
Yu, J.; Cheng, Y.; Zhao, R.-W.; Feng, R.; and Zhang, Y. 2022{\natexlab{a}}.
\newblock {MM-Pyramid}: Multimodal Pyramid Attentional Network for Audio-Visual Event Localization and Video Parsing.
\newblock In \emph{ACM MM}, 6241--6249.

\bibitem[{Yu et~al.(2022{\natexlab{b}})Yu, Cheng, Zhao, Feng, and Zhang}]{10.1145/3503161.3547869}
Yu, J.; Cheng, Y.; Zhao, R.-W.; Feng, R.; and Zhang, Y. 2022{\natexlab{b}}.
\newblock MM-Pyramid: Multimodal Pyramid Attentional Network for Audio-Visual Event Localization and Video Parsing.
\newblock In \emph{ACM MM}, 6241–6249.

\bibitem[{Yung-Hsuan~Lai(2023)}]{lai2023modality}
Yung-Hsuan~Lai, Y.-C. F.~W., Yen-Chun~Chen. 2023.
\newblock Modality-Independent Teachers Meet Weakly-Supervised Audio-Visual Event Parser.
\newblock In \emph{NeurIPS}.

\bibitem[{Zhang, Wu, and Li(2022)}]{zhang2022actionformer}
Zhang, C.-L.; Wu, J.; and Li, Y. 2022.
\newblock Actionformer: Localizing moments of actions with transformers.
\newblock In \emph{ECCV}, 492--510.

\bibitem[{Zhao, Thabet, and Ghanem(2021)}]{zhao2021video}
Zhao, C.; Thabet, A.~K.; and Ghanem, B. 2021.
\newblock Video self-stitching graph network for temporal action localization.
\newblock In \emph{ICCV}, 13658--13667.

\bibitem[{Zhao et~al.(2024)Zhao, Zhou, Guo, Zhao, and Chen}]{zhao2024mmcse}
Zhao, P.; Zhou, J.; Guo, D.; Zhao, Y.; and Chen, Y. 2024.
\newblock Multimodal Class-aware Semantic Enhancement Network for Audio-Visual Video Parsing.
\newblock \emph{arXiv preprint arXiv:2412.11248}.

\bibitem[{Zhou et~al.(2024{\natexlab{a}})Zhou, Guo, Guo, Mao, Hu, Zhong, Chang, and Wang}]{zhou2024towards}
Zhou, J.; Guo, D.; Guo, R.; Mao, Y.; Hu, J.; Zhong, Y.; Chang, X.; and Wang, M. 2024{\natexlab{a}}.
\newblock Towards Open-Vocabulary Audio-Visual Event Localization.
\newblock \emph{arXiv preprint arXiv:2411.11278}.

\bibitem[{Zhou et~al.(2024{\natexlab{b}})Zhou, Guo, Mao, Zhong, Chang, and Wang}]{zhou2024label}
Zhou, J.; Guo, D.; Mao, Y.; Zhong, Y.; Chang, X.; and Wang, M. 2024{\natexlab{b}}.
\newblock Label-anticipated Event Disentanglement for Audio-Visual Video Parsing.
\newblock In \emph{ECCV}, 1--22.

\bibitem[{Zhou, Guo, and Wang(2023)}]{zhou2023contrastive}
Zhou, J.; Guo, D.; and Wang, M. 2023.
\newblock Contrastive positive sample propagation along the audio-visual event line.
\newblock \emph{TPAMI}, 7239--7257.

\bibitem[{Zhou et~al.(2023)Zhou, Guo, Zhong, and Wang}]{zhou2023improving}
Zhou, J.; Guo, D.; Zhong, Y.; and Wang, M. 2023.
\newblock Improving audio-visual video parsing with pseudo visual labels.
\newblock \emph{arXiv preprint arXiv:2303.02344}.

\bibitem[{Zhou et~al.(2024{\natexlab{c}})Zhou, Guo, Zhong, and Wang}]{zhou2024vaplan}
Zhou, J.; Guo, D.; Zhong, Y.; and Wang, M. 2024{\natexlab{c}}.
\newblock Advancing Weakly-Supervised Audio-Visual Video Parsing via Segment-wise Pseudo Labeling.
\newblock \emph{IJCV}, 1--22.

\bibitem[{Zhou et~al.(2024{\natexlab{d}})Zhou, Shen, Wang, Zhang, Sun, Zhang, Birchfield, Guo, Kong, Wang et~al.}]{zhou2023avss}
Zhou, J.; Shen, X.; Wang, J.; Zhang, J.; Sun, W.; Zhang, J.; Birchfield, S.; Guo, D.; Kong, L.; Wang, M.; et~al. 2024{\natexlab{d}}.
\newblock Audio-Visual Segmentation with Semantics.
\newblock \emph{IJCV}, 1--21.

\bibitem[{Zhou et~al.(2022)Zhou, Wang, Zhang, Sun, Zhang, Birchfield, Guo, Kong, Wang, and Zhong}]{zhou2022avs}
Zhou, J.; Wang, J.; Zhang, J.; Sun, W.; Zhang, J.; Birchfield, S.; Guo, D.; Kong, L.; Wang, M.; and Zhong, Y. 2022.
\newblock Audio--visual segmentation.
\newblock In \emph{ECCV}, 386--403.

\bibitem[{Zhou et~al.(2021)Zhou, Zheng, Zhong, Hao, and Wang}]{zhou2021positive}
Zhou, J.; Zheng, L.; Zhong, Y.; Hao, S.; and Wang, M. 2021.
\newblock Positive sample propagation along the audio-visual event line.
\newblock In \emph{CVPR}, 8436--8444.

\end{thebibliography}

\newpage
\appendix
\section{Supplementary Material}
\subsection{A. Additional Ablation Studies on Core Modules}
The core modules in our proposed CCNet are the Cross-Modal Consistency Collaboration (CMCC) and the Multi-Temporal Granularity Collaboration (MTGC).
In this supplemental material, we first provide additional ablation studies on these modules.

\noindent\textbf{A.1. Ablation study on the modalities in CMCC module.}
We implement the CMCC module for both audio and visual modalities, leading to audio-guided CMCC and visual-guided CMCC (as shown in Fig. 2 of our main paper).
We investigate its effects on each modality.
As shown in Table~\ref{tab:ablation_CMCC_mod}, the model obtains the highest performance when the CMCC is utilized for both audio and visual modalities.
Performance decreases when applying CMCC to a single audio or visual modality, as expected. 
For example, the average mAP performance significantly decreases by 1.3\% when only employing CMCC for the audio modality.
These results indicate that it is vital for both modalities to learn from cross-modal consistency collaborations.

\begin{table}[h]
\setlength{\tabcolsep}{2mm}
      \resizebox{\linewidth}{!}{
  \begin{tabular}{cccccccc}
    \toprule
   \multicolumn{2}{c}{Modalities} & \multirow{2}{*}{0.5} & \multirow{2}{*}{0.6} & \multirow{2}{*}{0.7} &\multirow{2}{*}{0.8} &\multirow{2}{*}{0.9} &\multirow{2}{*}{Avg.}\\
    \cmidrule{1-2}
    audio & visual & & & & & & \\ \midrule
    \ding{52} & \ding{52} &\textbf{51.9} &\textbf{47.2} &\textbf{41.5} &\textbf{34.1} &\textbf{23.0} &\textbf{49.2}\\
    \ding{52} & \ding{56}  & 50.3 & 45.5 & 40.2 & 33.0 & 22.6 & 47.9\\
    \ding{56} & \ding{52}  & 50.5 & 45.7 & 40.4 & 33.4 & 22.9 & 47.9\\
    \bottomrule
\end{tabular}}
\caption{The ablation results of our CMCC module. 
  We examine the impacts of applying CMCC to audio and visual modalities.}
\label{tab:ablation_CMCC_mod}
\end{table}

\noindent\textbf{A.2. Ablation study on the number of granularity $L_c$ in CMCC module.}
We analyze model performance at different levels of granularity $L_c$. As shown in Table~\ref{tab:layer_number}, the model achieves the peak performance when $L_c=6$. 
When $L_c$ is increased or decreased, the model performance deteriorates.
Therefore, we select $L_c=6$ to implement our CMCC module.

\begin{table}[h]
\setlength{\tabcolsep}{3mm}
      \resizebox{\linewidth}{!}{
  \begin{tabular}{cccccccc}
    \toprule
    $L_c$ &0.5 & 0.6 &0.7 &0.8 &0.9 &Avg.\\
    \midrule
    4 & 50.7 & 44.6 & 37.7 & 29.3 & 18.7 & 47.3 \\
    5 & 52.1 & 46.4 & 41.0 & 32.9 & 21.0 & 48.7 \\
    \textbf{6}  & \textbf{51.9} & \textbf{43.6} & \textbf{38.5} & \textbf{34.1} & \textbf{23.0} & \textbf{49.2} \\
    7 & 51.3 & 46.5 & 40.5 & 33.7 & 22.7 & 48.4 \\
  \bottomrule
\end{tabular}}
 \caption{Ablation results on the number of granularity $L_c$ in CMCC module.}
\label{tab:layer_number}
\end{table}

\noindent\textbf{A.3. Comparison with modules from AVEL task.}
Our CMCC module is designed to effectively encode audio-visual relations, consisting of a cross-modal interaction branch and a temporal consistency-gated branch.
To further validate the superiority of the proposed CMCC module, we compare it with additional modules from a related Audio-Visual Event Localization (AVEL)~\cite{tian2018audio} task.
Specifically, we replace the CMCC module in our model with several representative modules from AVEL methods and train the adapted model on the UnAV-100 dataset.
The experimental results are presented in Table~\ref{tab:AVEL_result}.
Our CMCC-based model significantly outperforms the variants using modules from the AVEL task.
These comparison results further demonstrate the superiority of the proposed CMCC module.

\begin{table}[h]
\setlength{\tabcolsep}{1mm}
      \resizebox{\linewidth}{!}{
  \begin{tabular}{cccccccc}
    \toprule
    Methods &0.5 & 0.6 &0.7 &0.8 &0.9 &Avg.\\
    \midrule
    AVE~\cite{Tian_2018_ECCV} & 35.9 & 31.5 & 27.1 & 22.0 & 16.0 & 35.2 \\
    CMBS~\cite{9880210} & 45.9 & 40.3 & 33.9 & 27.9 & 18.3 & 43.7 \\
    PSP~\cite{zhou2021positive}  & 48.3 & 43.6 & 38.5 & 32.6 & 22.7 & 46.1 \\
    MM-Pyramid~\cite{10.1145/3503161.3547869} & 48.9 & 44.1 & 38.1 & 30.8 & 20.9 & 46.5 \\
    \textbf{CMCC-based (ours)} &\textbf{51.9} &\textbf{47.2} &\textbf{41.5} &\textbf{34.1} &\textbf{23.0} &\textbf{49.2} \\
  \bottomrule
\end{tabular}}
 \caption{Comparison between our CMCC module and other modules from a related AVEL~\cite{tian2018audio} task. We replaced the CMCC model in our model with those representative modules from the AVEL task and re-train it on the UnAV-100 dataset. }
\label{tab:AVEL_result}
\end{table}

\noindent\textbf{A.4. More ablation study of MTGC module.} The MTGC module is comprised by the Coarse-to-Fine Collaboration (C2F) and Fine-to-Coarse Collaboration (F2C) blocks.
We provide an additional ablation study on the MTGC module to explore the effects of different strategies for constructing fine-grained or coarse-grained temporal features.
For a feature at the $k$-th temporal granularity, the default C2F/F2C block configuration utilizes the temporal feature at the \textit{adjacent} granularity ($k$-1/$k$+1) as the corresponding fine-/coarse-grained feature.
Moreover, we investigate a variant of MTGC that concatenates features from \textit{all} granularities ($l_c < k$ or $l_c > k$) to serve as the fine- or {coarse}-grained features.
In Table~\ref{tab:coarse_feature_strategy_in_C2F}, we report the experimental results of these two strategies.
The results indicate that incorporating temporal clues from all granularities fails to confer additional advantages, implying that introducing features from broader ranges might lead to increased confusion for event predictions within multiple temporal durations.
Consequently, we opt for the `adjacent' strategy in implementing our MTGC module.

\begin{table}[h]
\setlength{\tabcolsep}{3mm}
      \resizebox{\linewidth}{!}{
  \begin{tabular}{cccccccc}
    \toprule
    Strategies &0.5 & 0.6 &0.7 &0.8 &0.9 &Avg.\\
    \midrule
    adj. &\textbf{51.9} &\textbf{47.2} &\textbf{41.5} &34.1 &23.0 &\textbf{49.2} \\
    all & 51.4 & 46.7 & \textbf{41.5} & \textbf{35.0} & \textbf{23.6} & 48.9 \\
  \bottomrule
\end{tabular}}
\caption{Different strategies for constructing coarse- or fine-grained temporal features in the MTGC module. For a given temporal feature at $k$-th granularity, the strategy `adj.' represents using only the feature at immediately adjacent granularity ($k$-1 for fine-grained; $k$+1 for coarse-grained), whereas the `all' strategy denotes employing features from all granularities ($l_c < k$ for fine-grained; $l_c > k$ for coarse-grained).}
\label{tab:coarse_feature_strategy_in_C2F}
\end{table}

\subsection{B. Evaluation on More Datasets}
In our main paper, we primarily evaluate our model on the UnAV-100~\cite{geng2023dense} dataset, which is also the only standard large-scale dataset released for the studied Dense Audio-Visual Event Localization (DAVEL) task.
To further validate the effectiveness and robustness of the proposed CCNet, we evaluate our model and compare it against prior methods on additional audio-visual datasets (originally proposed for other tasks), including ActivityNet1.3~\cite{Heilbron2015ActivityNetAL} and LFAV~\cite{10.1145/3672079}.

\noindent\textbf{B.1. Comparison on ActivityNet1.3 dataset.} 
The ActivityNet1.3 ~\cite{Heilbron2015ActivityNetAL} dataset is designed for the Temporal Action Localization (TAL) task, which contains 20k untrimmed audible videos spanning 200 classes of different actions.
Following the prior method UniAV~\cite{geng2024uniav}, we extract the audio and visual features using the pretrained ONE-PEACE~\cite{wang2023one} model and adopt the mean Average Precision (mAP) as the evaluation metric.
The experimental results are shown in Table \ref{tab:activity_result}.
Our model achieves an average mAP of 38.6\% and outperforms both methods from the TAL and DAVEL tasks by a large margin, demonstrating the superiority of the proposed method. 

\begin{table}[t]
\setlength{\tabcolsep}{3mm}
\resizebox{\linewidth}{!}{
  \begin{tabular}{ccccc}
    \toprule
    Methods & 0.5 & 0.75 & 0.95 & Avg. \\
    \midrule
    DAVEL~\cite{geng2023dense} & 50.5 & - & - & 32.5\\
    Tridet~\cite{Shi_2023_CVPR} &56.9 & - & - & 35.9\\
    UniAV~\cite{geng2024uniav} &56.8 &36.0 &\textbf{6.7} &36.2\\
    \textbf{CCNet (Ours)} & \textbf{60.1} & \textbf{39.6} & 6.2 & \textbf{38.6} \\
  \bottomrule
\end{tabular}}
 \caption{ Comparison results on the ActivityNet1.3 dataset. }
\label{tab:activity_result}
\end{table}

\noindent\textbf{B.2. Comparison on LFAV dataset.}  
We further evaluate our method on a related audio-visual dataset, LFAV~\cite{10.1145/3672079}. 
This dataset contains 5,175 untrimmed audible videos, providing frame-level annotations for audio events, visual events, and audio-visual events.
Notably, this dataset is designed for Long-form Audio-Visual Video Parsing (LAVVP), which differs from the studied DAVEL task.
Although both our method and the prior baseline DAVEL~\cite{geng2023dense} were not originally proposed for the LAVVP task, we can directly apply them to this dataset for further comparisons.
We adopt the official event-level F1 score as the evaluation metric.
As shown in Table~\ref{tab:LFAV_result}, our method continues to outperform the baseline DAVEL on this dataset, with a 1.82\% improvement in average performance and a 2.35\% improvement in audio-visual event localization.
These results further demonstrate the superiority of our method over the DAVEL method in temporally localizing events of varied durations.

\begin{table}[t]
\setlength{\tabcolsep}{1mm}
\resizebox{\linewidth}{!}{
  \begin{tabular}{ccccc}
    \toprule
    Methods & Audio & Visual & Audio-Visual & Avg. \\
    \midrule
    DAVEL~\cite{geng2023dense} & 22.58 & 29.20 & 21.79 & 24.52\\
    \textbf{CCNet (Ours)} & \textbf{23.64} & \textbf{31.24} & \textbf{24.14} & \textbf{26.34} \\
  \bottomrule
\end{tabular}}
 \caption{ Comparison results on the LFAV dataset.}
\label{tab:LFAV_result}
\end{table}

\end{document}